\documentclass{article}

\usepackage{PRIMEarxiv}

\usepackage[utf8]{inputenc} 
\usepackage[T1]{fontenc}    
\usepackage{hyperref}       
\usepackage{url}            
\usepackage{booktabs}       
\usepackage{amsfonts}       
\usepackage{nicefrac}       
\usepackage{microtype}      
\usepackage{lipsum}
\usepackage{fancyhdr}       
\usepackage{graphicx}       
\graphicspath{{media/}}     

\usepackage{algorithm}
\usepackage{algorithmic}
\usepackage{multirow}

\usepackage[normalem]{ulem}
\useunder{\uline}{\ul}{}
\usepackage{enumitem}
\usepackage{amsmath}
\usepackage{graphicx}
\usepackage{caption}

\usepackage{amsthm} 

\usepackage{amssymb}
\newtheorem{definition}{Definition}

\title{SGA: A Graph Augmentation Method for Signed Graph Neural Networks
}

\author{
  Zeyu Zhang \\
  Huazhong Agricultural University \\
  Wuhan, China \\
  \texttt{zhangzeyu@mail.hzau.edu.cn} \\
   \And   
  Shuyan Wan \\
  University of Electronic Science and Technology of China \\
  Chengdu, China \\
  \texttt{wanshuyanzzz@163.com} \\
     \And
  Sijie Wang, Xianda Zheng \\
  The University of Auckland \\
  Auckland, New Zealand \\
  \texttt{swan387, xzhe162@aucklanduni.ac.nz} \\
    \And
  Xinrui Zhang \\
  Southwest University \\
  Chongqing, China \\
  \texttt{zhangxinrui@email.swu.edu.cn} \\
    \And
  Kaiqi Zhao, Jiamou Liu \\
  The University of Auckland \\
  Auckland, New Zealand \\
  \texttt{jiamou.liu, kaiqi.zhao@auckland.ac.nz} \\
    \And   
  Dong Hao \\
  University of Electronic Science and Technology of China \\
  Chengdu, China \\
  \texttt{haodong@uestc.edu.cn} \\
}

\begin{document}
\maketitle

\begin{abstract}
Signed Graph Neural Networks (SGNNs) play a crucial role in the analysis of intricate patterns within real-world signed graphs, where both positive and negative links coexist. Nevertheless, there are three critical challenges in current signed graph representation learning using SGNNs. First, signed graphs exhibit significant sparsity, leaving numerous latent structures uncovered. Second, SGNN models encounter difficulties in deriving proper representations from unbalanced triangles. Finally, real-world signed graph datasets often lack supplementary information, such as node labels and node features. These challenges collectively constrain the representation learning potential of SGNN. We aim to address these issues through data augmentation techniques. However, the majority of graph data augmentation methods are designed for unsigned graphs, making them unsuitable for direct application to signed graphs. To the best of our knowledge, there are currently no data augmentation methods specifically tailored for signed graphs. In this paper, we propose a novel \underline{S}igned \underline{G}raph \underline{A}ugmentation framework, \textbf{SGA}. This framework primarily consists of three components. In the first part, we utilize the SGNN model to encode the signed graph, extracting latent structural information in the encoding space, which is then used as candidate augmentation structures. In the second part, we analyze these candidate samples (i.e., edges), selecting the most beneficial candidate edges to modify the original training set. In the third part, we introduce a new augmentation perspective, which assigns training samples different training difficulty, thus enabling the design of new training strategy. Extensive experiments on six real-world datasets, i.e., Bitcoin-alpha, Bitcoin-otc, Epinions, Slashdot, Wiki-elec and Wiki-RfA show that SGA improve the performance on multiple benchmarks. Our method outperforms baselines by up to 22.2\% in terms of AUC for SGCN on Wiki-RfA, 33.3\% in terms of F1-binary, 48.8\% in terms of F1-micro, and 36.3\% in terms of F1-macro for GAT on Bitcoin-alpha in link sign prediction. Our implementation is available in PyTorch\footnote{https://anonymous.4open.science/r/SGA-3127}.
\end{abstract}


\section{Introduction}
As social media continues to gain widespread popularity, it gives rise to a multitude of interactions among individuals, which are subsequently documented within social graphs \cite{leskovec2010predicting,kumar2018community}. While many of these social interactions denote positive connections, such as liking, trust, and friendship, there are also instances of negative interactions, encompassing feelings of hatred, distrust, and more. In essence, graphs that encompass both positive and negative interactions or links are commonly termed as {\em signed graphs} \cite{derr2020network,tang2016survey}. For instance, Slashdot \cite{leskovec2009community}, a tech-related news website, allows users to tag other users as either `friends' or `foes'. Such a situation can be naturally modeled as a {\em signed graph}. In recent years, there has been a growing interest among researchers in exploring network representation within the context of signed graphs \cite{wang2017signed,li2020learning,shu2021sgcl}. Most of these methods are combined with Graph Neural Networks (GNN) \cite{kipf2016semi,vaswani2017attention}, and are therefore collectively referred to as Signed Graph Neural Networks (SGNN) \cite{derr2018signed,huang2021sdgnn}. This endeavor is focused on acquiring low-dimensional representations of nodes, with the ultimate goal of facilitating subsequent network analysis tasks, especially {\em link sign prediction}. 

There are three issues within signed graph representation learning, i.e., 1) real-world signed graph datasets are exceptionally sparse \cite{tang2016survey}, with a significant amount of potential structure remaining uncollected or undiscovered, 2) according to the analysis in \cite{zhang2023rsgnn}, SGNN fails to learn proper representations from unbalanced triangles, despite the prevalence of unbalanced triangles in real-world datasets, 3) real-world signed graph datasets only contain structural information and lack more side information. Data augmentation which has been well-studied in computer vision \cite{devries2017improved,ho2019population,song2023deep,caron2020unsupervised} and natural language processing \cite{xia2019generalized,csahin2019data,fabbri2021improving} holds promise for alleviating the three issues mentioned above. 

\begin{figure}
    \centering
    \includegraphics[width=\columnwidth]{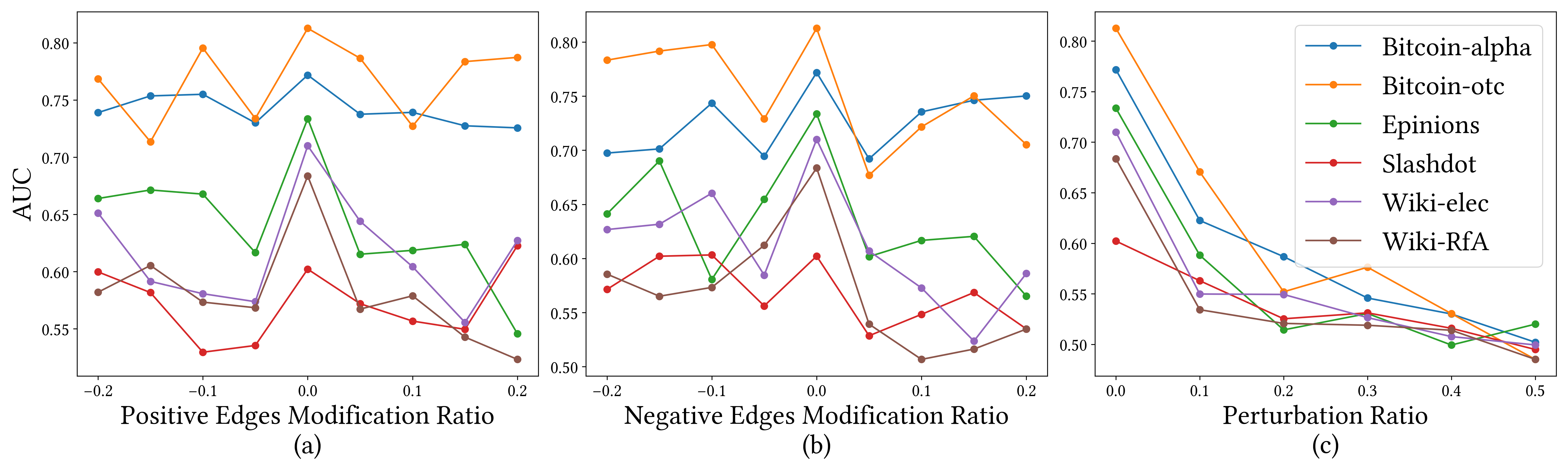}
    \caption{Effectiveness of data augmentation through random structural perturbations (SGCN \cite{derr2018signed} as backbone model) on link sign prediction performance. (a) Randomly increasing or decreasing positive edges. (b) Randomly increasing or decreasing negative edges. (c) Randomly flipping the sign of edges.}
    \label{fig:random}\vspace*{-0.2cm}
\end{figure}

In recent years, there have been significant advancements in graph data augmentation methods \cite{zhao2021data,han2022g,liu2022local}, including node perturbation \cite{you2020graph,huang2018adaptive}, edge perturbation \cite{rong2019dropedge} and sub-graph sampling \cite{wang2020graphcrop}. {\em Yet}, current graph data augmentation (GDA) methods cannot be directly applied to signed graphs. The limitation of current GDA methods primarily include the following three aspects: 1) Some graph augmentation methods \cite{zhao2021data,han2022g} incorporate side information such as node features and node labels. However, real-world signed graph datasets lack this kind of information and only possess structural information. 2) Random structural perturbation \cite{rong2019dropedge} is not applicable to signed graph neural networks. As shown in Figure \ref{fig:random}, we employed three different random methods (i.e., random addition/removal of positive edges, random addition/removal of negative edges, and random sign-flipping of existing edges) in conjunction with the classic SGNN model, SGCN \cite{derr2018signed}. The experimental results show that these random methods cannot enhance the performance of SGCN. The experiment results involving random edge flipping (Figure \ref{fig:random}(c)) demonstrate that the data augmentation methods employed in signed graph contrastive learning model \cite{shu2021sgcl,zhang2023contrastive} do not readily extend to general representation learning models. 3) Current data augmentation methods typically enhance data from aspects like node features and node labels, lacking novel augmentation perspectives \cite{ding2022data}.

To the best of our knowledge, there currently exists no data augmentation solution tailored specifically for SGNN models. In this paper, we embark on the exploration of data augmentation methods for signed graph representation learning. The primary challenges are as follows:

\begin{enumerate}
    \item Uncover potential structural information using only structural data.
    \item Design a more refined and targeted approach to modify the existing structure in order to mitigate the adverse effects of unbalanced cycles on SGNN models \cite{zhang2023rsgnn}.
    \item Propose a new data augmentation perspective specifically tailored for signed graphs.
\end{enumerate}

To address the aforementioned challenges, we propose a novel \underline{S}igned \underline{G}raph \underline{A}ugmentation framework, \textbf{SGA}. This framework primarily consists of three components which focuses on mining new structural information and edge features from the training samples (i.e., edges) with only structural information. In more details, to address the \textbf{first} challenge, we treat uncovering potential structural information as an edge prediction problem. We utilize a classic SGNN model, such as SGCN \cite{derr2018signed}, to encode the nodes of the signed graph. In the encoding space, We consider the relationships between nodes that are close in proximity as potential positive edges, while we view the relationships between nodes that are farther apart as potential negative edges. The newly discovered structural information is treated as candidate training samples (i.e., edges). To address the \textbf{second} challenge, We do not directly insert these candidate training samples into the training set but adopt a more cautious approach to discern whether they introduce harmful information. We demonstrate that only candidate training samples that do not decrease the local balance of nodes (see Def. \ref{def:local_balance_degree}) are beneficial candidate training samples. Based on this conclusion, we select the beneficial candidate training samples to insert into the training set. To address the \textbf{third} challenge, we introduce a new data augmentation perspective: {\em training difficulty} and assign different difficulty scores to various training samples. Differing from conventional training approaches that assign equal training weights to all samples, we have developed novel training schemes for SGNN models based on these varying difficulty levels. 

To evaluate the effectiveness of SGA, we perform extensive experiments on six real-world datasets, i.e., Bitcoin-alpha, Bitcoin-otc, Epinions, Slashdot, Wiki-elec and Wiki-RfA. We verify that our proposed framework SGA can enhance model performance using the common SGNN model SGCN\cite{derr2018signed} as the encoding module. The experimental results show that SGA improves the link sign prediction accuracy of five base models, including two unsigned GNN models (GCN\cite{kipf2016semi} and GAT\cite{velickovic2019deep}) and three signed GNN models (SGCN, SiGAT\cite{huang2019signed} and GS-GNN\cite{liu2021signed}). SGA boosts up to 22.2\% in terms of AUC for SGCN on Wiki-RfA, 33.3\% in terms of F1-binary, 48.8\% in terms of F1-micro, and 36.3\% in terms of F1-macro for GAT on Bitcoin-alpha in link sign prediction, at best. These experimental results demonstrate the effectiveness of SGA.

\begin{itemize}
    \item We are the first to introduce the research on data augmentation for signed graph neural networks.
    \item We have proposed a novel signed graph augmentation framework which tries to alleviating the three issues existing in signed graph neural networks. This framework not only helps uncover potential training samples but also aids in selecting beneficial samples to mitigate the introduction of harmful structural information. Additionally, it enables the augmentation of training samples with new feature (i.e., training difficulty see Def. \ref{def:difficulty_score}), which forms the basis for a new training strategy.
    \item Extensive experiments on six real-world datasets with five backbone models demonstrate the effectiveness of our framework.
\end{itemize}

\section{Related Work}
As described above, relevant topics about our paper is Signed Graph Neural Networks and Graph Augmentation. Next, we will discuss these two aspects separately.

\subsection{Signed Graph Neural Networks}

Due to the widespread popularity of social media, signed graphs have garnered significant attention in the field of network representation \cite{chen2018bridge, wang2018shine, zhang2023contrastive, zhang2023rsgnn}. Existing research has predominantly concentrated on tasks related to \textit{link sign prediction}, while overlooking other crucial tasks like node classification \cite{tang2016node}, node ranking \cite{jung2016personalized}, and community detection \cite{bonchi2019discovering}. Some signed graph embedding techniques, such as SNE \cite{yuan2017sne}, SIDE \cite{kim2018side}, SGDN \cite{jung2020signed}, and ROSE \cite{javari2020rose}, rely on random walks and linear probabilistic methods. In recent years, neural networks have been employed for signed graph representation learning. The first Signed Graph Neural Network (SGNN), SGCN \cite{derr2018signed}, generalizes GCN \cite{kipf2016semi} to signed graphs by utilizing balance theory to ascertain the positive and negative relationships between nodes separated by multiple hops. Another noteworthy GCN-based approach is GS-GNN, which relaxes the balance theory assumption and typically assumes nodes can be grouped into multiple categories. Additionally, prominent SGNN models like SiGAT \cite{huang2019signed}, SNEA \cite{li2020learning}, SDGNN \cite{huang2021sdgnn}, and SGCL \cite{shu2021sgcl} are based on GAT \cite{velivckovic2017graph}. These efforts mainly revolve around the development of more advanced SGNN models. Our work diverges from these approaches as we introduce a novel signed graph augmentation to improve the performance of SGNNs

\subsection{Graph Data Augmentation}
In response to the challenges posed by data noise and limited data availability in graph representation learning, there has been a recent surge in research focused on enhancing graph data augmentation techniques. \cite{park2021metropolis,han2022g,liu2022local}. According to a survey of graph data augmentation \cite{ding2022data}, graph augmentation methods can be classified into three types, i.e., feature-wise \cite{liu2022local,zhu2021graph,you2021graph}, structure-wise \cite{luo2021learning,xu2022graph,zheng2020robust} and label-wise \cite{zhang2017mixup,verma2021graphmix}. For feature-wise type, LAGNN \cite{liu2022local} enriches the node features by employing a generative model that takes as input the localized neighborhood information of the target node. Other feature-wise methods \cite{zhu2020deep,you2020graph} generate augmented node feature by random shuffling. Structure-wise augmentation methods target at modifying edges and nodes (e.g., randomly adding or deleting edges). GAUG \cite{zhao2021data} employs neural edge predictors can effectively encode class-homophilic structure to promote intra-class edges and demote inter-class edges in given graph structure. GraphSMOTE \cite{zhao2021graphsmote} insert nodes to enrich the minority classes. Graph diffusion method (GDC \cite{gasteiger2019diffusion}) can generate an augmented graph by providing the global views of the underlying structure. Label-wise augmentation methods aim at augmenting the limited labeled training data. G-Mixup \cite{han2022g} augment graphs for graph classification by interpolating the generator (i.e., graphon) of different classes of graphs. It is worth noting that most of these data augmentation methods rely on additional information such as node features and node labels. However, for signed graphs, these types of information are absent, making these methods not directly applicable to data augmentation in signed networks.

\section{Problem Statement}

A {\em signed graph} is defined as $\mathcal{G} = (\mathcal{V}, \mathcal{E}^{+}, \mathcal{E}^{-})$, where $\mathcal{V} = \{v_1, \ldots, v_{|\mathcal{V}|}\}$ represents the set of nodes, and $\mathcal{E}^+$ and $\mathcal{E}^-$ denote the positive and negative edges, respectively. Each edge $e_{ij} \in \mathcal{E}^{+} \cup \mathcal{E}^{-}$ connecting two nodes $v_i$ and $v_j$ can be either positive or negative, but not both, meaning that $\mathcal{E}^{+} \cap \mathcal{E}^{-} = \varnothing$. We use $\sigma(e_{ij}) \in \{+,-\}$ to denote the {\em sign} of $e_{ij}$. The structure of $\mathcal{G}$ is represented by the adjacency matrix $A \in \mathbb{R}^{|\mathcal{V}| \times |\mathcal{V}|}$, where each entry $A_{ij} \in \{1,-1,0\}$ signifies the sign of the edge $e_{ij}$. It's important to note that, unlike unsigned graph datasets, signed graphs typically do not provide node features, meaning there is no feature vector $x_i$ associated with each node $v_i$.

{\em Positive} and {\em negative neighbors} of $v_i$ are denoted as $\mathcal{N}_{i}^{+}=\{v_j\mid A_{ij}>0\}$ and $\mathcal{N}_{i}^{-}=\{v_j\mid A_{ij}<0\}$, respectively. Let $\mathcal{N}_i=\mathcal{N}_i^+\cup \mathcal{N}_i^-$ be the set of neighbors of node $v_i$. 
$\mathcal{O}_3$ represents the set of {\em triangles} in the signed graph, i.e., $\mathcal{O}_3=\{\{v_i,v_j,v_k\}\mid A_{ij}A_{jk}A_{ik}\neq 0\}$. A triangle $\{v_i,v_j,v_k\}$ is called {\em balanced} if $A_{ij}A_{jk}A_{ik}>0$, and is called {\em unbalanced} otherwise. 

\textbf{Problem Definition}: $D_{\text{train}} \cup D_{\text{test}} = \mathcal{E}^{+} \cup \mathcal{E}^{-}$, where $D_{\text{train}}$ refers to the set of train samples (edges) and $D_{\text{test}}$ refers to the set of test samples. When only given $D_{\text{train}}$, our purpose is to design a graph augmentation strategy $\psi : (D_{\text{train}}) \rightarrow (D^{\prime}_{\text{train}}, \mathcal{F})$, where $D^{\prime}_{\text{train}}$ refers to augmented train edge set and $\mathcal{F}$ refers to the newly generated edge features.

\section{Proposed Method}

\begin{figure*}[t]
    \centering
    \includegraphics[width=\columnwidth]{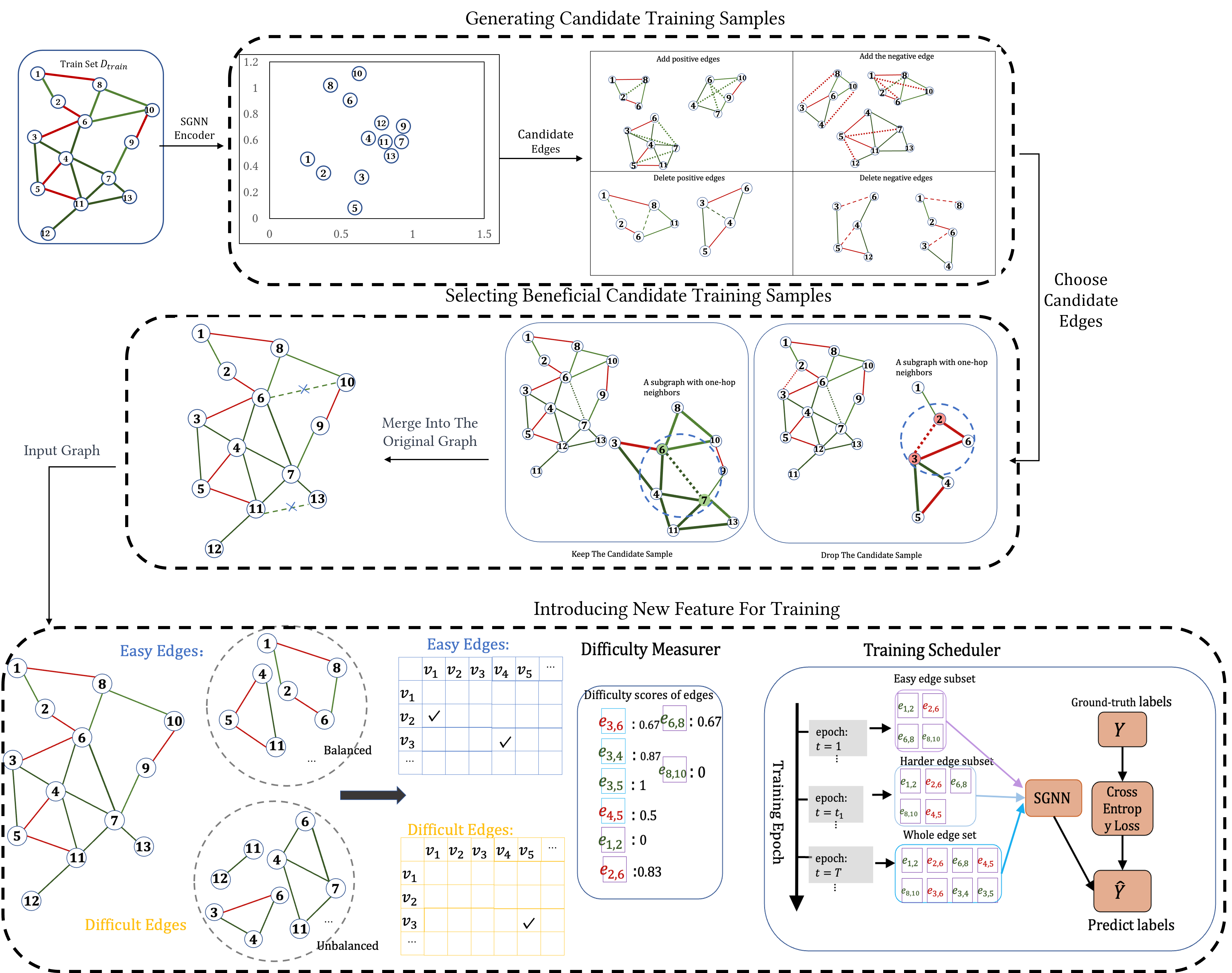}
    \caption{The overall architecture of SGA. Green lines represent positive edges and red lines represent negative edges.}
    \label{fig:framework}
\end{figure*}

In this section, we present a \underline{S}igned \underline{G}raph \underline{A}ugmentation framework, aiming to augment training samples (i.e., edges) from structure perspective (edge manipulation) to side information (edge feature). Figure shows the overall architecture. SGA encompasses three key elements: 1) generating new training candidate samples, 2) Selecting beneficial training samples, and 3) introducing new feature (training difficulty) for training samples. To be specific, we \textbf{first} utilize the SGNN model for encoding the nodes within the signed graph. Within this encoding space, our objective is to unearth latent relationships between nodes and produce fresh training candidate samples, specifically edges. On an intuitive level, we posit that nodes in close proximity within the encoding space are inclined to form positive relationships (positive edges), whereas nodes further apart are more likely to establish negative relationships (negative edges). \textbf{Subsequently}, we conduct a theoretical analysis, highlighting that only training samples that do not decrease the local balance of nodes (see Def. \ref{def:local_balance_degree}) are beneficial candidate training samples. \textbf{Lastly}, we propose a new graph augmentation perspective, assign a difficulty score to each training sample and use this feature to guide the training process. Intuitively, we aim for the backbone model to prioritize the retention of structural information with lower difficulty scores while downplaying the significance of structural information with higher difficulty scores.

\subsection{Generating Candidate Training Samples}\label{sec:Generating Candidate Training Samples}

Real-world signed graph datasets are extremely sparse, with many missing or uncollected relationships between nodes. In this subsection, we attempt to uncover the potential relationships between nodes. We first use SGNN model, e.g., SGCN \cite{derr2018signed}, the classical SGNN model, as the encoder to project nodes from topological space to embedding space. Here, the node representations are updated by aggregating information from different types of neighbors as follows:

For the first aggregation layer $\ell=1$:

\begin{equation}
\begin{aligned}
    H^{pos(1)} &= \sigma \left(\mathbf{W}^{pos(1)} \left[A^{+}H^{(0)}, H^{(0)} \right]\right) \\
    H^{neg(1)} &= \sigma \left(\mathbf{W}^{neg(1)} \left[A^{-}H^{(0)}, H^{(0)} \right]\right) \\
\end{aligned}
\end{equation}

For the aggregation layer $\ell>1$:

\begin{equation}
\begin{aligned}
    H^{pos(\ell)} &= \sigma \left( \mathbf{W}^{pos(\ell)} \left[A^{+}H^{pos(\ell-1)}, A^{-}H^{neg(\ell-1)}, H^{pos(\ell-1)} \right]\right) \\
    H^{neg(\ell)} &= \sigma \left( \mathbf{W}^{neg(\ell)} \left[A^{+}H^{neg(\ell-1)}, A^{-}H^{pos(\ell-1)}, H^{neg(\ell-1)} \right]\right), \\
\end{aligned}
\end{equation}
where $H^{pos(\ell)} (H^{neg(\ell)})$ are positive (negative) part of representation matrix at the $\ell$th layer. $A^{+} (A^{-})$ are the row normalized matrix of positive (negative) part of the adjacency matrix $A$. $\mathbf{W}^{pos(\ell)} (\mathbf{W}^{neg(\ell)})$ are learnable parameters of positive (negative) part, and $\sigma(\cdot)$ is the activation function. $\left[.\right]$ is the concatenation operation. After conducting message-passing for $L$ layers, the final node representation matrix is $Z = H^{(L)} = \left[H^{pos(L)},H^{neg(L)} \right]$. For node $v_i$, the node embedding is $Z_i$. As we wish to classify whether a pair of node are with a positive, negative or no edge between them. We train a multinomial logistic regression classifier (MLG) \cite{derr2018signed}. The training loss is as follows:

\begin{equation}
\footnotesize
\mathcal{L}\left(\theta^{\text{MLG}}\right)= -\frac{1}{|D_{\text{train}}|} \sum_{\left(v_i, v_j, \sigma(e_{ij})\right) \in D_{\text{train}}}  \log \frac{\exp \left(\left[Z_i, Z_j\right] \theta_{\sigma(e_{ij})}^{\text{MLG}}\right)}{\Sigma_{q \in\{+,-, ?\}} \exp \left(\left[Z_i, Z_j\right] \theta_q^{\text{MLG}}\right)}
\end{equation}

$\theta^{\text{MLG}}$ refers to the parameter of the MLG classifier. Using this classifier, for any two node $v_i$, $v_j$ we can calculate the probability of forming a positive or negative edge between any two nodes, denoted as $Pr_{e_{ij}}^{pos}$ and $Pr_{e_{ij}}^{neg}$. We configure four probability threshold hyper-parameters, i.e., the probability threshold for adding positive edges ($\epsilon^{+}_{add}$), the probability threshold for adding negative edges ($\epsilon^{-}_{add}$), the probability threshold for deleting positive edges ($\epsilon^{+}_{del}$), the probability threshold for deleting negative edges ($\epsilon^{-}_{del}$),  We adopt the following strategy to generate candidate training samples:

\begin{itemize}
    \item $\forall v_i, v_j \in \mathcal{V}$, if $\left(v_i, v_j, \sigma(e_{ij})\right) \notin D_{\text{train}}$, $Pr_{e_{ij}}^{pos} > \epsilon^{+}_{add} \vee Pr_{e_{ij}}^{neg} > \epsilon^{-}_{add}$, then $D_{\text{train\_add}}^{\text{cand.}} \cup \left(v_i, v_j, \sigma(e_{ij})\right)$
    \item $\forall v_i, v_j \in \mathcal{V}$, if $\left(v_i, v_j, \sigma(e_{ij})\right) \in D_{\text{train}}$, $A_{ij}>0$, $Pr_{e_{ij}}^{pos} < \epsilon^{+}_{del}$, then $D_{\text{train\_del}}^{\text{cand.}} \cup \left(v_i, v_j, \sigma(e_{ij})\right)$
    \item $\forall v_i, v_j \in \mathcal{V}$, if $\left(v_i, v_j, \sigma(e_{ij})\right) \in D_{\text{train}}$, $A_{ij}<0$, $Pr_{e_{ij}}^{neg} < \epsilon^{-}_{del}$, then $D_{\text{train\_del}}^{\text{cand.}} \cup \left(v_i, v_j, \sigma(e_{ij})\right)$
\end{itemize}

$D_{\text{train\_add}}^{\text{cand.}}$ and $D_{\text{train\_del}}^{\text{cand.}}$ respectively refer to the candidate training set for adding edges and the candidate training set for deleting edges.

\subsection{Selecting Beneficial Candidate Training Samples}\label{sec:Selecting Beneficial Candidate Training Samples}
After generating candidate training set $D_{\text{train\_add}}^{\text{cand.}}$ and $D_{\text{train\_del}}^{\text{cand.}}$, we next aim to incorporate these training candidates into the training sample set $D_{\text{train}}$. One issue that needs to be addressed is that not all new generated training samples (i.e., edges) result in positive effects. Hence, we need to select benefical candidate training samples.

\begin{figure}
    \centering
    \includegraphics[width=0.5\columnwidth]{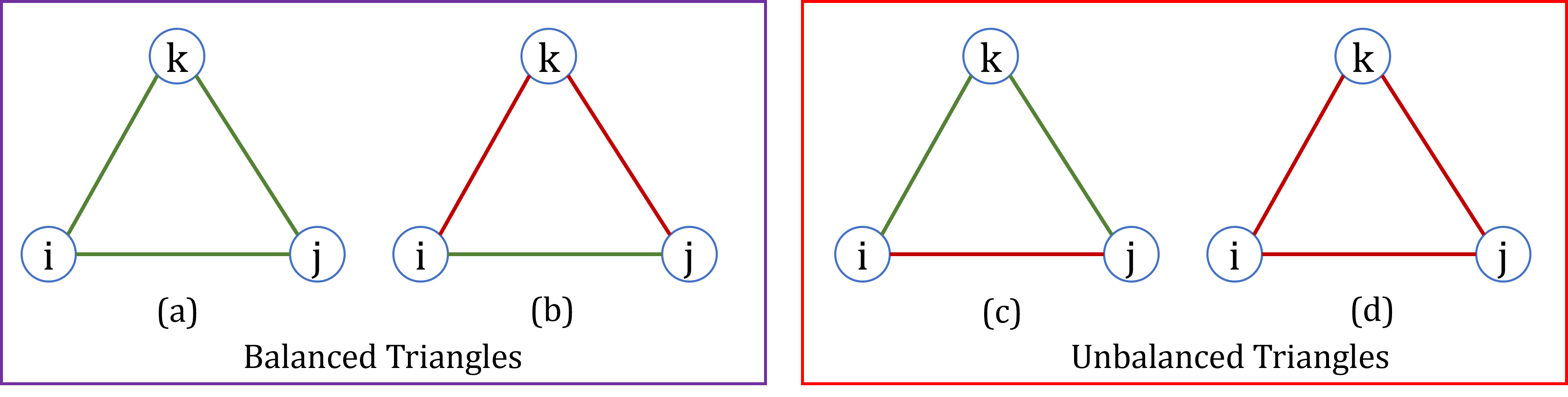}
    \caption{Four isomorphism types of triangles. Green and red lines represent positive and negative edges, resp.}
    \label{fig:triad}\vspace*{-0.2cm}
\end{figure}

According to \cite{zhang2023rsgnn,zhang2023contrastive}, SGNN models that rely on balance theory cannot learn a proper representation for nodes from unbalanced triangles, as is shown in Figure \ref{fig:triad}. 

\begin{definition}\label{def:triangles}
{\em Balanced (unbalanced) triangles} are cycles with 3 nodes containing even (odd) negative edges.
\end{definition}

The addition of both positive and negative edges can result in changes in the local structure of nodes, potentially leading to unbalanced triangles. We provide a definition of local balance degree.

\begin{definition}[Local Balance Degree]\label{def:local_balance_degree}
For node $v_{i}$, the local balance degree is defined by:
\begin{equation}
\label{eq:local_degree}
D_3(v_{i})=\frac{|\mathcal{O}_3^{+}(v_{i})| - |\mathcal{O}_3^{-}(v_{i})|}{|\mathcal{O}_3^{+}(v_{i})| + |\mathcal{O}_3^{-}(v_{i})|}
\end{equation}
where $\mathcal{O}_3^{+}(v_{i})$ ($\mathcal{O}_3^{-}(v_{i})$) represents the set of balanced (unbalanced) triangles containing node $v_{i}$. $|\cdot|$ represents the set cardinal number.
\end{definition}

From Def. \ref{def:local_balance_degree}, we can observe that the node's local balance degree is related to the count of balanced and unbalanced triangles that include this node. Based on this definition, we can conclude that beneficial candidate training samples does not decrease the local balance degree of target nodes.

\subsection{Introducing New feature for Training Samples}\label{sec:Introducing New feature for Training Samples}
\begin{figure}
    \centering
    \includegraphics[width=0.5\columnwidth]{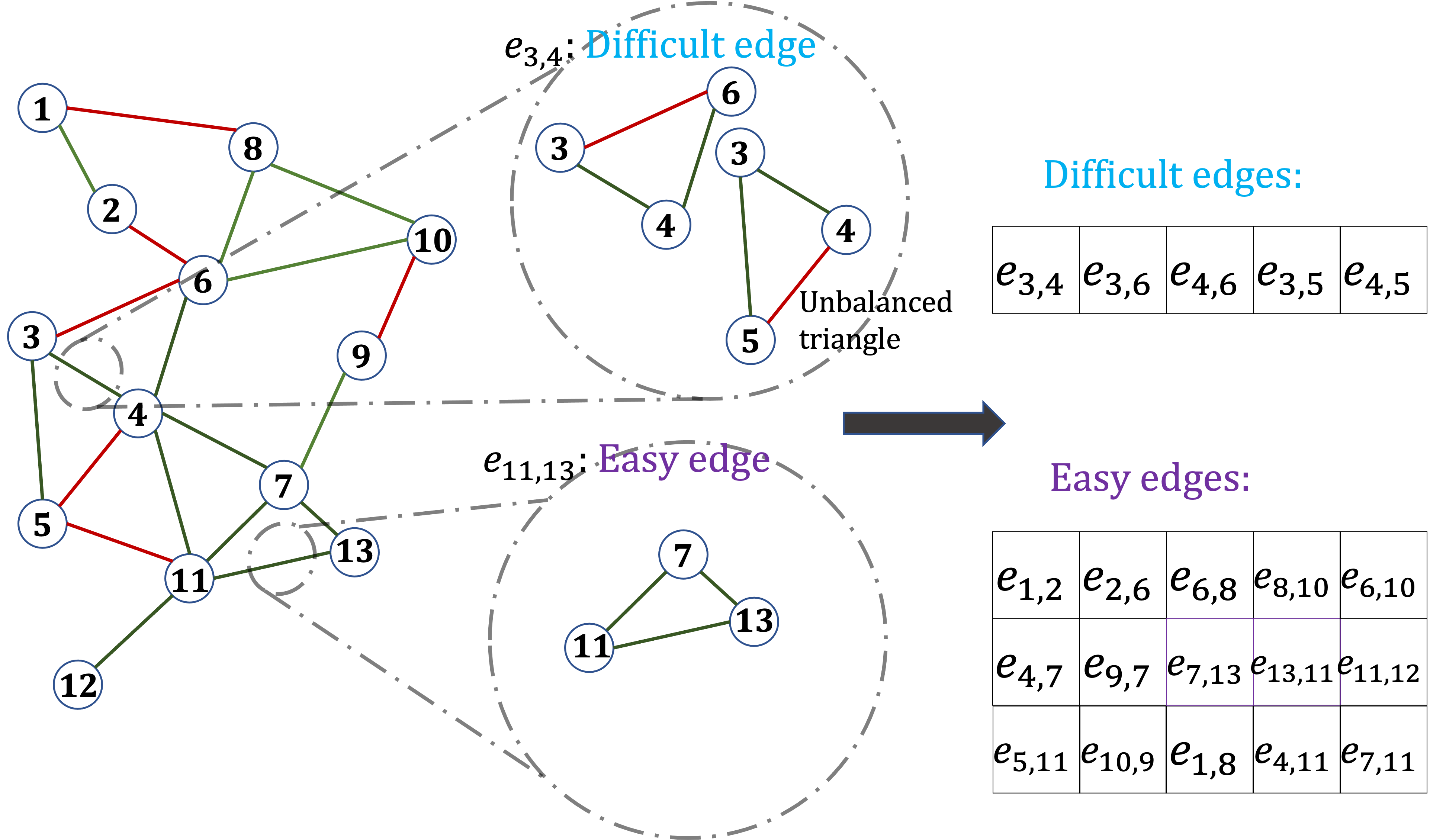}
    \caption{Illustration of node difficulty, where green lines represent positive edges and red lines represent negative edges.}
    \label{fig:difficulty}\vspace*{-0.2cm}
\end{figure}

Based on the above analysis, it is apparent that unbalanced triangles pose a challenging task for SGNNs. Intuitively, when an edge is part of an unbalanced triangle, its level of difficulty in terms of representation learning should surpass that of edges not involved in such structures, as shown in Figure \ref{fig:difficulty}. 

\begin{definition}[Edge Difficulty Score]\label{def:difficulty_score}
    For edge $e_{ij}$, the difficulty score is defined by:
\begin{equation}
    \text{Score}(e_{ij}) = 1 - \frac{D_3(v_{i}) + D_3(v_{j})}{2}
\end{equation}
where $D_3(v_{i}$ and $D_3(v_{j})$ refer to the local balance degree of node $v_i$ and $v_j$, respectively.
\end{definition}

Upon quantifying the difficulty scores for each edge within the training set, a curriculum-based training approach is applied to enhance the performance of the SGNN model. This curriculum is fashioned following the principles set forth in \cite{wei2023clnode}, which enables the creation of a structured progression from easy to difficult. The process entails initially sorting the training set $\mathcal{E}$ in ascending order based on their respective difficulty scores. Subsequently, a pacing function $g(t)$ is employed to allocate these edges to distinct training epochs, transitioning from easier to more challenging instances, where $t$ signifies the $t$-th epoch.we use linear pacing function as shown below:

\section{Experiments}
In this section, we commence by assessing the enhancements brought about by SGA in comparison to diverse backbone models for the link sign prediction task. We will answer the following questions:

\begin{itemize}[leftmargin=0.5cm]
    \item \textbf{Q1}: Can SGA framework increase the performance of backbone models?
    \item \textbf{Q2}: Do each part of the SGA framework play a positive role?
    \item \textbf{Q3}: Is the proposed method sensitive to hyper-parameters ? How do key hyper-parameters impact the method performance?
\end{itemize}

\subsection{Datasets}
We conduct experiments on six real-world datasets, i.e., Bitcoin-OTC, Bitcoin-Alpha, Wiki-elec, Wiki-RfA, Epinions, and Slashdot. The main statistics of each dataset are summarized in Table \ref{tab:datasets}. In the following, we explain important characteristics of the datasets briefly.

\textbf{Bitcoin-OTC}\footnote{http://www.bitcoin-otc.com} \cite{kumar2016edge,kumar2018rev2} and \textbf{Bitcoin-Alpha}\footnote{http://www.btc-alpha.com} are two datasets extracted from bitcoin trading platforms. Due to the fact Bitcoin accounts are anonymous, people give trust or not-trust tags to others in order to enhance security.

\textbf{Wiki-elec}\footnote{https://www.wikipedia.org} \cite{leskovec2010signed,leskovec2010predicting} is a voting network in which users can choose trust or distrust to other users in administer elections. \textbf{Wiki-RfA} \cite{west2014exploiting} is a more recent version of Wiki-elec.

\textbf{Epinions}\footnote{http://www.epinions.com} \cite{leskovec2010signed} is a consumer review site with trust and distrust relationships between users.

\textbf{Slashdot}\footnote{http://www.slashdot.com} \cite{leskovec2010signed} is a technology-related news website in which users can tag each other as friends (trust) or enemies (distrust).

\begin{table}[t]
\centering
\caption{The statistics of datasets.}
\begin{tabular}{cccc}
\hline
Dataset       & \# Links & \# Positive Links & \# Negative Links \\ \hline
Bitcoin-OTC   & 35,592    & 32,029             & 3,563              \\
Bitcoin-Alpha & 24.186    & 22,650             & 1,536              \\
Wiki-elec      & 103,689   & 81,345             & 22,344             \\
Wiki-RfA       & 170,335   & 133,330            & 37,005             \\
Epinions      & 840,799   & 717,129            & 123,670            \\
Slashdot      & 549,202   & 425,072            & 124,130            \\ \hline
\end{tabular}
\label{tab:datasets}
\end{table}

Following the experimental settings in \cite{derr2018signed}, We randomly split the edges into a training set and a testing set with a ratio 8:2. We run with different train-test splits for 5 times to get the average scores and standard deviation.

\subsection{Baselines and Experiment Setting}
We use five popular graph representation learning models as backbones, including both unsigned GNN models and signed GNN models.

\textbf{Unsigned GNN}: We employ two classical GNN models (i.e., GCN \cite{kipf2016semi} and GAT \cite{velickovic2019deep}). These methods are designed for unsigned graphs, thus, as mentioned before, we consider all edges as positive edges to learn node embeddings in the experiments.

\textbf{Signed Graph Neural Networks}: SGCN \cite{derr2018signed} and SiGAT~\cite{huang2019signed} respectively generalize GCN~\cite{kipf2016semi} and GAT \cite{velickovic2019deep} to signed graphs based on message mechanism. Besides, SGCN integrates the balance theory. GS-GNN \cite{liu2021signed} adopts a more generalized assumption (than balance theory) that nodes can be divided into multiple latent groups. We use these signed graph representation models as baselines to explore whether \textbf{SGA} can enhance their performance. 

We implement our SGA using PyTorch \cite{paszke2019pytorch} and employ PyTorch Geometric \cite{Fey/Lenssen/2019} as its complementary graph library. The graph encoder, responsible for augmenting the graph, consists of a 2-layer SGCN with an embedding dimension of 64. This encoder is optimized using the Adam optimizer, set with a learning rate of 0.01 over 300 epochs. To ensure a consistent comparison, we randomly standardized the node embedding dimension at 64 across all embedding-based methods, matching the dimensionality used in GS-GNN \cite{huang2021signed}. For the baseline methods, we adhere to the parameter configurations as recommended in their originating papers. Specifically, for unsigned baseline models like GCN and GAT, we employ the Adam optimizer, with a learning rate of 1e-2, a weight decay of 5e-4, and span the training over 500 epochs. In contrast, signed baseline models are trained with an initial learning rate of 5e-3, a weight decay of 1e-5, and are run for 3000 epochs.

The experiments were performed on a Linux machine with eight 24GB NVIDIA GeForce RTX 3090 GPUs.

Our primary evaluation criterion is link sign prediction—a binary classification task. We assess the performance employing AUC, F1-binary, F1-macro, and F1-micro metrics, consistent with the established norms in related literature~\cite{huang2021signed,liu2021signed}. It's imperative to note that across these evaluation metrics, a higher score directly translates to enhanced model performance.

\subsection{Performance on Link Sign Prediction (Q1)}

To comprehensively evaluate the performance of our proposed SGA, we contrast it with several baseline configurations that exclude SGA integration. For a detailed view, AUC and F1-binary score results are presented in Table \ref{tab:main_res}. Further, the F1-micro and F1-macro scores can be referenced in Table \ref{tab:f1mima}. For each model, the mean AUC and F1-binary scores, along with their respective standard deviations, are documented. These metrics are derived from five independent runs on each dataset, utilizing distinct, non-overlapping splits: 80\% of the links are earmarked for training, while the residual 20\% serve as the test set. Additionally, the table elucidates the percentage improvement in these metrics attributable to the integration of SGA, relative to the baseline models devoid of SGA. The results proffer several salient insights:

\begin{itemize}[leftmargin=0.5cm]
    \item Our investigations affirm that the SGA framework serves as a potent catalyst in augmenting the performance of both signed and unsigned graph neural networks. This underscores the efficacy of tailoring the inherent graph structure and the subsequent training regimen, enabling models to astutely discern intricate node relationships.
    \item Within the realm of unsigned graph neural networks, the GAT model exhibits a more pronounced enhancement in performance relative to the GCN in the majority of scenarios. This observation is attributable to GAT's comparatively modest baseline performance in relation to GCN, engendering a larger margin for refinement. This phenomenon accentuates the potential of the SGA framework to stabilize the GAT's training dynamics by refining the graph topology and the associated training procedure.
    \item Among the signed neural networks we scrutinized—SGCN, SiGAT, and GS-GNN—it's pivotal to note that only SGCN integrates the balance theory. This strategic incorporation propels SGCN to manifest the most pronounced improvements across a majority of the datasets. Even though SGCN's performance tends to be lackluster when leveraging random embeddings for initial node representations, our findings suggest that aligning the dataset more closely with the model's intrinsic assumptions can pave the way for superior performance outcomes.
    \item A salient advantage conferred by the SGA framework is the enhanced stability observed in signed graph neural network models. This stability is palpable through reduced standard deviation in AUC and F1-binary scores across most datasets. Contrarily, unsigned baseline models, which inherently overlook the nuanced negative inter-node relationships, do not seem to reap similar stability dividends.
\end{itemize}

\begin{table}[t]
\centering
\caption{Link sign prediction results (average $\pm$ standard deviation) with AUC (\%) and F1-binary (\%) on six benchmark datasets.}
\label{tab:main_res}
\scriptsize
\begin{tabular}{c|cc|cc|cc|cc|cc|cc}
\hline
Datasets & \multicolumn{2}{c|}{Bitcoin-alpha} & \multicolumn{2}{c|}{Bitcoin-otc} & \multicolumn{2}{c|}{Epinions} & \multicolumn{2}{c|}{Slashdot} & \multicolumn{2}{c|}{Wiki-elec} & \multicolumn{2}{c}{Wiki-RfA} \\
Methods & AUC & F1-binary & AUC & F1-binary & AUC & F1-binary & AUC & F1-binary & AUC & F1-binary & AUC & F1-binary \\ \hline
GCN & 60.9$\pm$0.8 & 73.6$\pm$1.5 & 69.1$\pm$0.8 & 83.0$\pm$1.5 & 68.5$\pm$0.2 & 80.4$\pm$0.2 & 51.8$\pm$0.7 & 55.8$\pm$1.9 & 64.0$\pm$1.1 & 75.5$\pm$1.5 & 60.4$\pm$0.7 & 72.0$\pm$1.0 \\
+SGA & 64.5$\pm$1.4 & 80.7$\pm$2.5 & 69.3$\pm$1.0 & 89.0$\pm$0.9 & 69.3$\pm$0.2 & 82.5$\pm$0.4 & 51.5$\pm$2.0 & 61.9$\pm$10.1 & 64.8$\pm$0.3 & 75.0$\pm$1.7 & 60.9$\pm$0.3 & 72.9$\pm$1.6 \\
(Improv.) & 5.9\% & 9.7\% & 0.3\% & 7.2\% & 1.2\% & 2.6\% & - & 10.9\% & 1.3\% & - & 0.8\% & 1.3\% \\ \hline
GAT & 59.3$\pm$1.6 & 68.0$\pm$19.3 & 67.2$\pm$2.1 & 84.6$\pm$5.9 & 53.1$\pm$1.7 & 68.0$\pm$19.0 & 51.4$\pm$1.6 & 61.0$\pm$18.5 & 54.6$\pm$2.3 & 69.8$\pm$15.1 & 51.9$\pm$1.1 & 72.8$\pm$4.5 \\
+SGA & 65.0$\pm$6.8 & 90.7$\pm$3.1 & 74.3$\pm$2.0 & 93.6$\pm$0.5 & 51.9$\pm$2.1 & 73.0$\pm$16.9 & 58.9$\pm$4.5 & 70.9$\pm$12.9 & 56.8$\pm$4.7 & 73.3$\pm$6.1 & 59.7$\pm$5.7 & 72.6$\pm$17.5 \\
(Improv.) & 9.6\% & 33.3\% & 10.6\% & 10.6\% & - & 7.4\% & 14.6\% & 16.2\% & 4\% & 5\% & 15\% & - \\ \hline
SGCN & 75.3$\pm$0.2 & 90.5$\pm$0.8 & 79.4$\pm$1.5 & 92.3$\pm$1.2 & 68.6$\pm$4.4 & 90.5$\pm$1.4 & 61.0$\pm$1.6 & 67.3$\pm$3.3 & 66.9$\pm$3.7 & 77.4$\pm$8.2 & 62.2$\pm$6.9 & 71.3$\pm$3.9 \\
+SGA & 80.6$\pm$1.2 & 93.9$\pm$0.4 & 82.2$\pm$0.7 & 94.7$\pm$0.7 & 78.1$\pm$0.4 & 92.8$\pm$0.4 & 63.2$\pm$1.2 & 84.6$\pm$0.8 & 77.7$\pm$0.2 & 87.0$\pm$0.7 & 76.0$\pm$0.5 & 86.9$\pm$0.4 \\
(Improv.) & 7.1\% & 3.8\% & 3.4\% & 2.6\% & 14\% & 2.6\% & 3.6\% & 25.7\% & 16.1\% & 12.3\% & 22.2\% & 21.8\% \\ \hline
SiGAT & 79.7$\pm$4.0 & 96.7$\pm$0.4 & 87.7$\pm$0.8 & 95.2$\pm$0.2 & 82.8$\pm$4.3 & 93.4$\pm$0.7 & 80.6$\pm$2.1 & 86.7$\pm$2.1 & 87.0$\pm$0.3 & 90.3$\pm$0.1 & 87.1$\pm$0.1 & 90.3$\pm$0.0 \\
+SGA & 87.4$\pm$0.8 & 96.3$\pm$0.3 & 89.8$\pm$0.9 & 95.1$\pm$0.4 & 88.4$\pm$1.5 & 94.6$\pm$0.6 & 85.0$\pm$0.6 & 89.3$\pm$0.3 & 88.5$\pm$0.3 & 90.5$\pm$0.2 & 88.0$\pm$0.2 & 90.3$\pm$0.1 \\
(Improv.) & 9.7\% & - & 2.4\% & - & 6.8\% & 1.3\% & 5.5\% & 3\% & 1.6\% & 0.3\% & 1\% & - \\ \hline
GSGNN & 85.1$\pm$1.3 & 97.0$\pm$0.1 & 88.3$\pm$1.1 & 95.9$\pm$0.3 & 88.9$\pm$0.4 & 95.0$\pm$0.6 & 77.9$\pm$0.7 & 88.6$\pm$0.3 & 88.2$\pm$0.2 & 90.9$\pm$0.1 & 86.8$\pm$0.2 & 90.3$\pm$0.2 \\
+SGA & 89.9$\pm$0.7 & 96.4$\pm$0.2 & 90.7$\pm$0.9 & 96.0$\pm$0.2 & 90.1$\pm$0.3 & 95.1$\pm$0.4 & 81.3$\pm$0.5 & 88.0$\pm$0.5 & 89.1$\pm$0.1 & 91.1$\pm$0.1 & 87.6$\pm$0.2 & 90.5$\pm$0.1 \\
(Improv.) & 5.7\% & - & 2.8\% & 0.1\% & 1.4\% & 0.1\% & 4.3\% & - & 1\% & 0.1\% & 0.9\% & 0.3\% \\ \hline
\end{tabular}
\end{table}

\begin{table}[t]
\centering
\caption{Link sign prediction results (average $\pm$ standard deviation) with F1-micro (\%) and F1-macro (\%) on six benchmark datasets.}
\scriptsize
\label{tab:f1mima}
\begin{tabular}{c|cc|cc|cc|cc|cc|cc}
\hline
Datasets & \multicolumn{2}{c|}{Bitcoin-alpha} & \multicolumn{2}{c|}{Bitcoin-otc} & \multicolumn{2}{c|}{Epinions} & \multicolumn{2}{c|}{Slashdot} & \multicolumn{2}{c|}{Wiki-elec} & \multicolumn{2}{c}{Wiki-RfA} \\
Methods & F1-micro & F1-macro & F1-micro & F1-macro & F1-micro & F1-macro & F1-micro & F1-macro & F1-micro & F1-macro & F1-micro & F1-macro \\ \hline
GCN & 59.9$\pm$1.8 &  45.0$\pm$0.9 & 72.9$\pm$2.1 & 57.7$\pm$1.4 & 70.4$\pm$0.2 &   60.0$\pm$0.2 & 47.1$\pm$1.4 &  44.9$\pm$1.0 & 65.8$\pm$1.5 & 59.5$\pm$1.1 &  61.8$\pm$1.0 & 55.9$\pm$0.8 \\
+SGA & 68.9$\pm$3.3 & 50.2$\pm$1.9 & 81.3$\pm$1.3 & 62.9$\pm$1.1 &  73.0$\pm$0.6 & 61.8$\pm$0.4 & 52.1$\pm$8.4 & 46.2$\pm$1.4 & 65.5$\pm$1.7 & 59.6$\pm$0.9 & 62.7$\pm$1.5 & 56.5$\pm$0.7 \\
(Improv.)  & 15.0\% & 11.6\% & 11.5\% & 9.0\% & 3.7\% & 3.0\% & 10.6\% & 2.9\% & - & 0.2\% & 1.5\% & 1.1\% \\ \hline
GAT & 56.2$\pm$20.7 & 41.9$\pm$10.3 & 75.2$\pm$8.1 & 58.5$\pm$4.9 & 58.1$\pm$19.8 & 44.8$\pm$6.3 & 53.1$\pm$15.2 &  44.5$\pm$5.0 &  60.2$\pm$13.0 & 50.7$\pm$6.1 & 60.8$\pm$4.9 & 50.6$\pm$1.7 \\
+SGA & 83.6$\pm$5.1 & 57.1$\pm$1.9 & 88.7$\pm$0.9 & 71.8$\pm$1.5 & 62.5$\pm$17.7 & 46.5$\pm$5.8 & 62.1$\pm$11.3 & 53.3$\pm$5.4 & 62.4$\pm$6.8 & 54.3$\pm$5.3 & 64.2$\pm$15.2 &  56.0$\pm$10.8 \\
(Improv.)  & 48.8\% & 36.3\% & 18.0\% & 22.7\% & 7.6\% & 3.8\% & 16.9\% & 19.8\% & 3.7\% & 7.1\% & 5.6\% & 10.7\% \\ \hline
SGCN & 83.4$\pm$1.3 & 62.1$\pm$1.1 & 86.7$\pm$1.9 & 72.0$\pm$1.5 & 83.9$\pm$2.0 & 68.0$\pm$2.9 & 58.2$\pm$3.1 & 54.6$\pm$2.4 & 69.2$\pm$7.6 & 61.8$\pm$3.9 & 61.7$\pm$4.1 & 56.3$\pm$4.6 \\
+SGA & 89.2$\pm$0.6 & 69.7$\pm$0.5 & 90.6$\pm$1.1 & 77.6$\pm$1.7 & 87.9$\pm$0.6 & 76.9$\pm$0.8 & 75.8$\pm$1.0 & 63.8$\pm$1.2 & 80.6$\pm$0.8 & 74.3$\pm$0.5 & 80.2$\pm$0.5 & 73.5$\pm$0.5 \\
(Improv.)  & 6.9\% & 12.2\% & 4.5\% & 7.8\% & 4.8\% & 13.0\% & 3.6\% & 25.7\% & 16.1\% & 12.3\% & 22.2\% & 21.8\% \\ \hline
SiGAT & 93.7$\pm$0.7 & 62.0$\pm$3.9 & 91.2$\pm$0.3 & 68.5$\pm$1.8 & 88.2$\pm$1.5 & 67.8$\pm$9.7 & 79.1$\pm$2.3 & 68.5$\pm$1.2 & 84.1$\pm$0.3 & 73.3$\pm$1.0 & 84.3$\pm$0.1 & 74.4$\pm$0.3 \\
+SGA & 93.0$\pm$0.5 & 69.9$\pm$0.9 & 91.2$\pm$0.7 & 76.5$\pm$1.7 & 90.7$\pm$0.9 & 79.9$\pm$1.4 & 82.9$\pm$0.5 & 73.7$\pm$0.8 & 85.0$\pm$0.3 & 77.0$\pm$0.1 & 84.6$\pm$0.2 & 75.9$\pm$0.4 \\
(Improv.)  & - & 12.7\% & 0.7\% & 11.7\% & 2.8\% & 17.8\% & 4.8\% & 7.6\% & 1.1\% & 5.0\% & 0.3\% & 2.0\% \\ \hline
GS-GNN & 94.2$\pm$0.3 & 62.3$\pm$10.1 & 93.1$\pm$1.0 & 75.4$\pm$7.5 & 91.7$\pm$1.0 & 81.9$\pm$1.7 & 81.6$\pm$0.4 & 70.3$\pm$1.0 & 85.3$\pm$0.1 & 75.8$\pm$0.6 & 84.1$\pm$0.3 & 73.0$\pm$1.1 \\
+SGA & 93.2$\pm$0.4 & 73.0$\pm$2.9 & 92.9$\pm$0.3 & 80.5$\pm$1.0 & 91.6$\pm$0.7 & 82.1$\pm$1.0 & 80.4$\pm$1.2 & 66.2$\pm$5.9 & 85.5$\pm$0.1 & 76.4$\pm$0.2 & 84.6$\pm$0.1 & 74.7$\pm$0.3 \\
(Improv.)  & - & 17.1\% & - & 6.8\% & - & 0.2\% & - & - & 0.3\% & 0.8\% & 0.7\% & 2.3\% \\ \hline
\end{tabular}
\end{table}

\subsection{Ablation Study (Q2)}
To ascertain the contributions of various components of SGA towards the model's overall performance, we systematically dissect and evaluate the SGCN under different conditions. Below are the configurations we investigate:

\begin{itemize}[leftmargin=0.5cm]
    \item \textbf{SGCN:} This configuration deploys SGCN on the original graph, devoid of any curriculum learning integration.
    \item \textbf{+SA (Structure Augmentation, refer to Sec. \ref{sec:Generating Candidate Training Samples} and Sec. \ref{sec:Selecting Beneficial Candidate Training Samples}):} SGCN operates on augmented datasets. This augmentation involves the addition or removal of edges from the initial graph.
    \item \textbf{+TP (Training Plan, refer to Sec. \ref{sec:Introducing New feature for Training Samples}):} SGCN runs on the original graph, but with a modified training paradigm. Adopting a curriculum learning approach, we rank edges by their "difficulty". The model is then progressively exposed to these edges, transitioning from simpler to more challenging ones as training epochs progress.
    \item \textbf{+SGA:} This is a holistic approach where SGCN operates on an augmented graph and also incorporates the aforementioned training plan. The model learns easier edges initially, gradually advancing to more intricate ones.
\end{itemize}

Our meticulous ablation study, encapsulated in Table \ref{tab:ablation} and executed across six benchmark datasets, bequeaths a panoply of profound revelations:
\begin{itemize}[leftmargin=0.5cm]
    \item \textbf{Significance of Data Augmentation:} Data Augmentation emerges as a pivotal component in enhancing the model's efficacy. Even when employed in isolation, it frequently outperforms the baseline model trained on the original graph without any specialized training strategy.
    \item \textbf{Isolated Impact of the Training Plan:} Introducing the Training Plan, devoid of concurrent modifications, occasionally offers subtle performance augmentations vis-à-vis Data Augmentation alone. An illustrative case is its behavior on the Bitcoin-alpha dataset: the Training Process modality engenders a decrement in AUC yet an ascendant trajectory in F1 score metrics.
    \item \textbf{Synergistic Benefits of Training Plan and Data Augmentation:} Melding the Training Plan with Data Augmentation orchestrates a synergy, frequently magnifying the individual benefits of Data Augmentation. A notable anomaly is discerned within the Slashdot dataset metrics: Data Augmentation in isolation showcases an elevated F1 score albeit a diminished AUC, in contrast even to the pristine SGCN technique. However, when the Training Plan is integrated, it champions superior AUC outcomes. Concurrently, the SGA strategy harmoniously amalgamates these attributes, delivering a commendable equilibrium of AUC and F1 scores, underscoring a symbiotic interplay between the methodologies.
\end{itemize}

\begin{table}[t]
\centering
\caption{The ablation study results of using different component of SGA.}
\label{tab:ablation}
\begin{tabular}{cc|cccc}
\hline
Dataset                   & Metric   & SGCN         & +SA                   & +TP                   & +SGA                  \\ \hline
\multirow{4}{*}{Bitcoin-alpha} & AUC & 75.3$\pm$0.2 & {\ul 78.4}$\pm$0.3 & 73.8$\pm$1.2 & \textbf{80.6}$\pm$1.0 \\
                          & F1-binary       & 90.5$\pm$0.8 & {\ul 93.3}$\pm$0.3    & 91.5$\pm$0.1          & \textbf{94.0}$\pm$0.3 \\
                          & F1-micro & 83.4$\pm$1.3 & {\ul 88.1}$\pm$0.4    & 84.9$\pm$0.1          & \textbf{89.2}$\pm$0.6 \\
                          & F1-macro & 62.1$\pm$1.1 & {\ul 67.5}$\pm$0.4    & 62.7$\pm$0.4          & \textbf{69.7}$\pm$0.5 \\ \hline
\multirow{4}{*}{Bitcoin-otc}   & AUC & 79.4$\pm$1.5 & {\ul 80.8}$\pm$1.4 & 79.7$\pm$0.9 & \textbf{82.2}$\pm$0.7 \\
                          & F1-binary       & 92.3$\pm$1.2 & {\ul 94.7}$\pm$0.9    & 92.3$\pm$1.5          & \textbf{94.7}$\pm$0.7 \\
                          & F1-micro & 86.7$\pm$1.9 & {\ul 90.7}$\pm$1.6    & 86.7$\pm$2.4          & \textbf{90.6}$\pm$1.1 \\
                          & F1-macro & 72.0$\pm$1.5 & {\ul 77.3}$\pm$2.3    & 72.1$\pm$2.8          & \textbf{77.6}$\pm$1.7 \\ \hline
\multirow{4}{*}{Epinions} & AUC      & 68.6$\pm$4.4 & {\ul 75.5}$\pm$2.2    & 75.0$\pm$0.5          & \textbf{78.1}$\pm$0.4 \\
                          & F1-binary       & 90.5$\pm$1.4 & {\ul 91.6}$\pm$0.3    & 88.6$\pm$2.6          & \textbf{92.8}$\pm$0.4 \\
                          & F1-micro & 83.9$\pm$2.0 & {\ul 85.9}$\pm$0.5    & 81.7$\pm$3.5          & \textbf{87.9}$\pm$0.6 \\
                          & F1-macro & 68.0$\pm$2.9 & {\ul 73.7}$\pm$1.3    & 70.1$\pm$2.5          & \textbf{76.9}$\pm$0.8 \\ \hline
\multirow{4}{*}{Slashdot} & AUC      & 61.0$\pm$1.6 & 58.8$\pm$5.1          & \textbf{63.3}$\pm$0.3 & {\ul 63.2}$\pm$1.2    \\
                          & F1-binary       & 67.3$\pm$3.3 & \textbf{85.7}$\pm$1.1 & 69.2$\pm$6.4          & {\ul 84.6}$\pm$0.8    \\
                          & F1-micro & 58.2$\pm$3.1 & \textbf{76.6}$\pm$1.1 & 60.7$\pm$6.0          & {\ul 75.8}$\pm$1.0    \\
                          & F1-macro & 54.6$\pm$2.4 & {\ul 58.2}$\pm$7.6    & 56.6$\pm$3.2          & \textbf{63.8}$\pm$1.2 \\ \hline
\multirow{4}{*}{Wiki-elec}     & AUC & 66.9$\pm$3.7 & {\ul 77.5}$\pm$0.4 & 70.4$\pm$2.6 & \textbf{77.7}$\pm$0.2 \\
                          & F1-binary       & 77.4$\pm$8.2 & {\ul 86.7}$\pm$0.7    & 74.5$\pm$9.2          & \textbf{87.0}$\pm$0.7 \\
                          & F1-micro & 69.2$\pm$7.6 & {\ul 80.3}$\pm$0.8    & 67.2$\pm$8.1          & \textbf{80.6}$\pm$0.8 \\
                          & F1-macro & 61.8$\pm$3.9 & {\ul 74.1}$\pm$0.5    & 62.3$\pm$5.1          & \textbf{74.3}$\pm$0.5 \\ \hline
\multirow{4}{*}{Wiki-RfA} & AUC      & 62.2$\pm$6.9 & {\ul 75.6}$\pm$0.7    & 68.6$\pm$3.7          & \textbf{76.0}$\pm$0.5 \\
                          & F1-binary       & 71.3$\pm$3.9 & {\ul 86.1}$\pm$1.1    & 75.7$\pm$4.3          & \textbf{86.9}$\pm$0.4 \\
                          & F1-micro & 61.7$\pm$4.1 & {\ul 79.4}$\pm$1.3    & 67.2$\pm$3.9          & \textbf{80.2}$\pm$0.5 \\
                          & F1-macro & 56.3$\pm$4.6 & {\ul 72.9}$\pm$1.0    & 62.0$\pm$3.0          & \textbf{73.5}$\pm$0.5 \\ \hline
\end{tabular}
\end{table}

\subsection{Parameter Sensitivity Analysis (Q3)}
\begin{figure}[t]
    \centering
    \includegraphics[width=1\columnwidth]{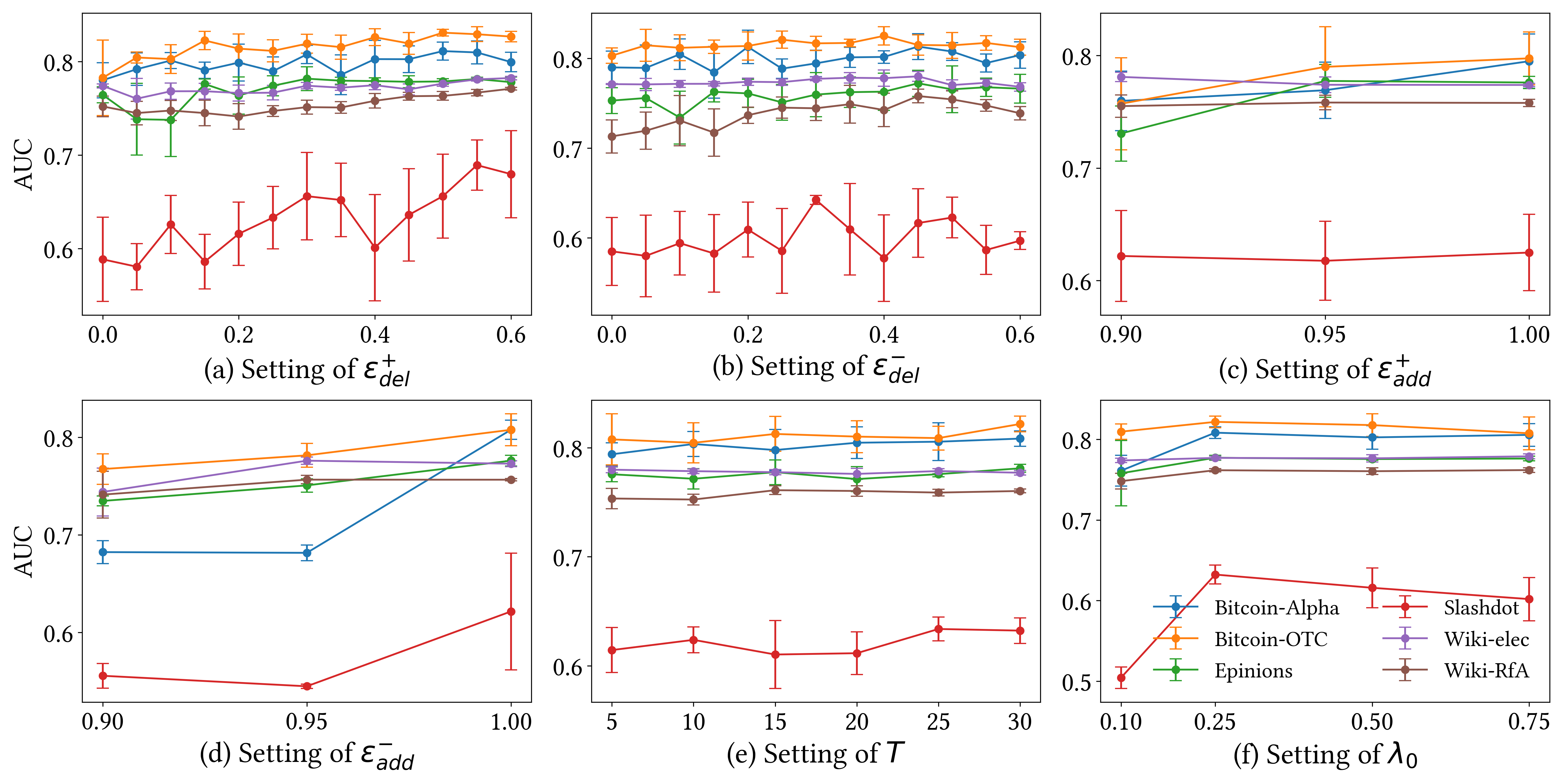}
    \caption{Performance of SGCN: AUC scores (with standard deviation) across six benchmark datasets, evaluated under variations in parameters $\epsilon_{del}^+$, $\epsilon_{del}^-$, $\epsilon_{add}^+$, $\epsilon_{add}^-$, $T$ and $\lambda_0$.}\label{fig:sens_auc}
\end{figure}

\begin{figure}[t]
    \centering
    \includegraphics[width=1\columnwidth]{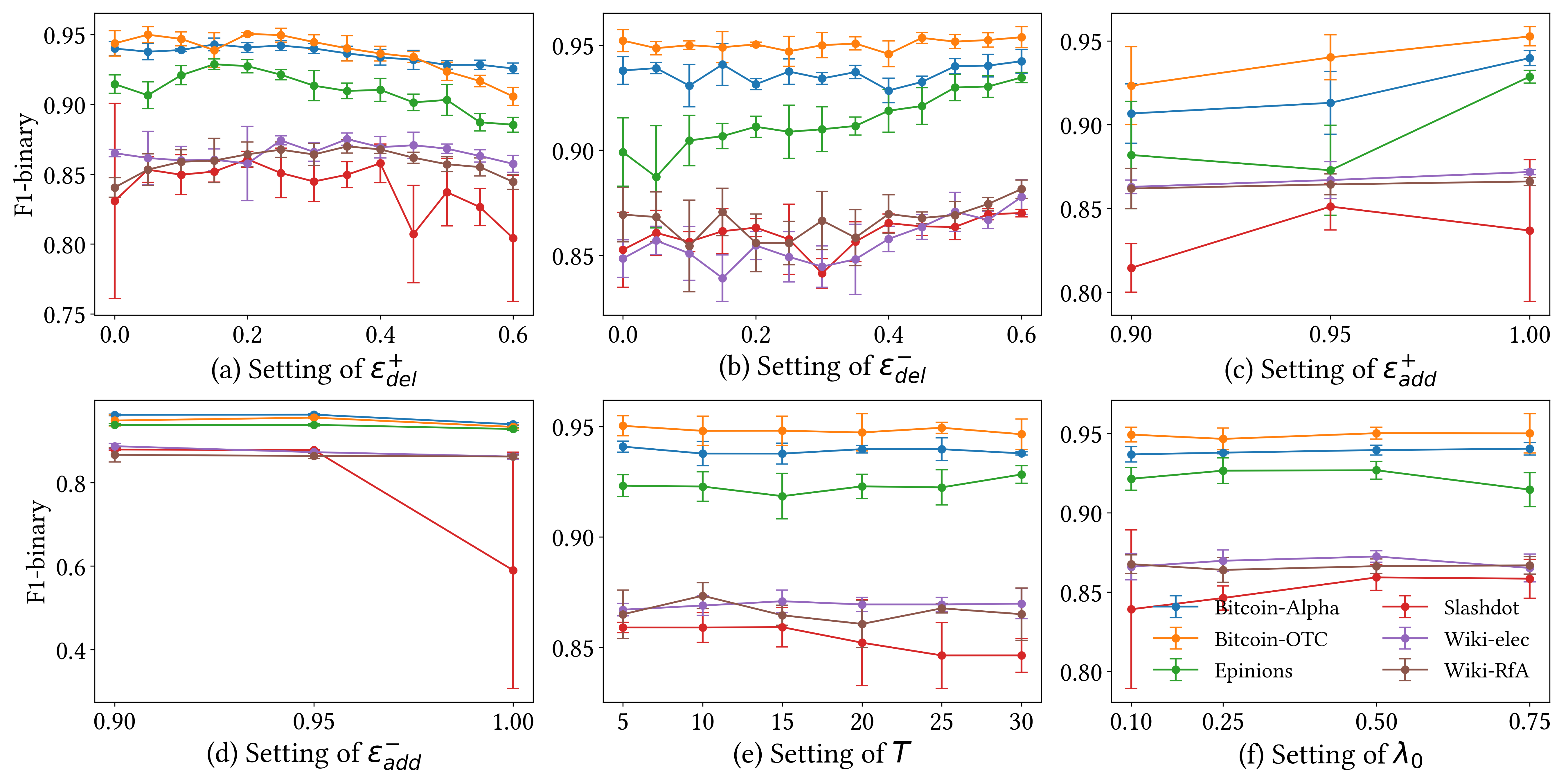}
    \caption{Performance of SGCN: F1-binary scores (with standard deviation) across six benchmark datasets, evaluated under variations in parameters $\epsilon_{del}^+$, $\epsilon_{del}^-$, $\epsilon_{add}^+$, $\epsilon_{add}^-$, $T$ and $\lambda_0$.}\label{fig:sens_f1}
\end{figure}

In this subsection, we undertake a sensitivity analysis focusing on six hyper-parameters: $\epsilon_{del}^+$, $\epsilon_{del}^-$, $\epsilon_{add}^+$, $\epsilon_{add}^-$ (these delineate the probability thresholds for adding or removing positive/negative edges); $T$ represents the number of intervals during the training process where more challenging edges are incrementally added to the training set; and $\lambda_0$ designates the initial fraction of the easiest examples. Performance metrics for the SGCN model within the SGA framework, as measured by AUC and F1-binary score across various hyper-parameter configurations, are illustrated in Figures \ref{fig:sens_auc} and \ref{fig:sens_f1} respectively.

From the figures, it's evident that the patterns of AUC and F1 score diverge depending on the adjustments to the hyper-parameters. Further, the SGCN performance on distinct datasets, like Slashdot, displays more pronounced variance in both AUC and F1 score relative to other datasets. On a broader scale, the AUC is fairly consistent with changes to $\epsilon_{del}^-$, $\epsilon_{add}^+$, and $T$. Notably, as $\epsilon_{del}^+$ or $\epsilon_{add}^-$ values rise, there's a tendency for the AUC to augment. Interestingly, AUC initially increases and then undergoes a minor dip as $\lambda_0$ rises. Regarding the F1 score, its sensitivity is limited concerning changes to $\epsilon_{add}^-$, $T$, and $\lambda_0$—with the Slashdot dataset being an outlier. In general, an increase in $\epsilon_{del}^-$ and $\epsilon_{add}^+$ boosts the F1 score. However, for $\epsilon_{del}^+$, the optimal value can differ across datasets, typically lying between 0.1 and 0.3.

\section{Conclusion}
Data augmentation is a widely embraced technique for enhancing the generalization of machine learning models, but its application to Graph Neural Networks (GNNs) presents distinctive challenges due to graph irregularity. Our groundbreaking research focuses on data augmentation for signed graph neural networks, introducing a novel framework to alleviate three significant issues in this field: exceptionally sparse real-world signed graph datasets, the difficulty in learning from unbalanced triangles, and the absence of side information in these datasets. In this paper, we propose a novel Signed Graph Augmentation framework, SGA. This framework is primarily composed of three key components. In summary, this framework has three main components: encoding structural information using the SGNN model, selecting beneficial edges for modification, and introducing a new perspective on training difficulty for improved training strategies. Through extensive experiments on benchmark datasets, our Signed Graph Augmentation (SGA) framework proves its versatility in boosting various SGNN models. As a promising future direction, further exploration of the theoretical foundations of signed graph augmentation is warranted, alongside the development of more potent methods to analyze diverse downstream tasks on signed graphs.

\bibliographystyle{unsrt}  
\bibliography{references}

\end{document}